\documentclass[conference]{IEEEtran}
\IEEEoverridecommandlockouts

\usepackage{hyperref}
\usepackage{amsmath,amssymb,amsfonts}
\usepackage{graphicx}
\usepackage{textcomp}
\usepackage{algorithm}
\usepackage{algpseudocode}
\usepackage{multirow}
\usepackage{comment}
\usepackage{graphicx}
\usepackage{amsmath}
\usepackage{cite}
\usepackage{comment}
\usepackage{subcaption}
\usepackage{algpseudocode}
\usepackage{amsmath}
\usepackage{comment}
\usepackage{multirow}
\usepackage{booktabs}
\usepackage{balance} 
\usepackage{array}
\usepackage[utf8]{inputenc}
\usepackage{xcolor}
\usepackage{url}
\usepackage{xurl}

\usepackage{authblk}  
\usepackage{float} 
\usepackage{cite}
\usepackage{enumitem}
\usepackage[normalem]{ulem}
\def\BibTeX{{\rm B\kern-.05em{\sc i\kern-.025em b}\kern-.08em
    T\kern-.1667em\lower.7ex\hbox{E}\kern-.125emX}}
\begin{document}

\title{Test-Time Adaptive Composition for Machine Learning as a Service (MLaaS) in IoT Environments}

\renewcommand\Authsep{, }
\renewcommand\Authand{, }
\renewcommand\Authands{, }
\setlength{\affilsep}{0.3em}

\author[]{Deepak Kanneganti}
\author[]{Sajib Mistry}
\author[]{Sheik Mohammad Mostakim Fattah}
\author[]{Aneesh Krishna}

\affil[]{\textit{School of EECMS, Curtin University, Australia}}
\affil[]{\{s.kanneganti, sajib.mistry, sheik.fattah, a.krishna\}@curtin.edu.au}

\maketitle



\begin{abstract}



The dynamic nature of Internet of Things (IoT) environments affects the long-term effectiveness of Machine Learning as a Service (MLaaS) compositions. Existing adaptive composition methods are mainly based on service replacement or re-composition, where identifying suitable substitutes is difficult and time-consuming. To address this, we propose a novel Test-Time Adaptive (TTA) composition framework for MLaaS in IoT environments. First, we introduce a TTA-aware composability model to determine whether adapted services remain compatible with the existing composition. Next, we design a service-level adaptation model to adjust individual services during inference while preserving composition performance. Experimental results demonstrate that the proposed framework reduces computational time more effectively than traditional adaptive approaches.

\end{abstract}

\newcounter{sagemakerfn}

\section{Introduction}

\IEEEPARstart{I}{n} recent years, Machine Learning as a Service (MLaaS) has become a key component of modern cloud and IoT ecosystems. It offers scalable tools to train, deploy, and manage ML models~\cite{ribeiro2015mlaas}. Major cloud providers offer MLaaS through platforms such as \textit{Microsoft Azure}\footnote{\url{https://azure.microsoft.com/en-us/products/machine-learning/}}, \textit{Amazon SageMaker}\footnote{\url{https://aws.amazon.com/sagemaker/}}, and the \textit{OpenAI ChatGPT}\footnote{\url{https://platform.openai.com/docs/}}. IoT environments increasingly rely on MLaaS to leverage AI-driven capabilities across domains such as healthcare and smart city\cite{kanneganti2024hybrid}. For example, \textit{Medtronic}, a healthcare IoT, integrates \textit{IBM IQcast} MLaaS to predict real-time low blood sugar events~\cite{medtronic2019sugariq}.

IoT environments often have complex service requirements such as \textit{diverse functionality}, \textit{high accuracy}, and \textit{low latency}. A \textit{single} MLaaS service may not always fulfill such requirements~\cite{fadlullah2018delay}. MLaaS composition enables IoT environments to integrate multiple MLaaS services to jointly meet complex \textit{functional} and \textit{QoS} requirements. For example, an IoT environment may require a human activity recognition (HAR) service for different users (e.g., varying medical conditions) and different sensor types (e.g., wearable and ambient sensors)~\cite{wang2025collaborative}. To address this, IoT environments compose multiple MLaaS services, resulting in a more robust HAR solution.

IoT environments are \textit{inherently dynamic} due to variations in user behaviour, data distributions, and system requirements, which affect composition performance~\cite{fadlullah2018delay}. Existing studies on service composition often address this challenge using adaptive approaches based on service substitution or re-composition techniques~\cite{kanneganti2025adaptive}. However, identifying suitable substitutes is challenging and time-consuming, particularly in the context of MLaaS~\cite{urbieta2017adaptive}. Moreover, substituted services may not always deliver the expected performance~\cite{fattah2019long}.

An effective alternative to substitution or re-composition is to enable personalization at the service composition level. Applying personalization at this level allows MLaaS workflows to dynamically adapt to heterogeneous user requirements and data characteristics\cite{ibrahim2015semantic}. This improves overall performance and relevance beyond what individual services can achieve. MLaaS services inherently support personalization to accommodate evolving user requirements. To the best of our knowledge, personalization of MLaaS services at the composition level remains largely unexplored. Therefore, this paper focuses on enabling \textit{personalized adaptation in MLaaS composition}.

To enable personalized adaptation at the composition level, we leverage Test-Time Adaptation (TTA), an emerging paradigm in ML for handling domain and data distribution shifts after model deployment~\cite{wang2020tent}. Models trained on historical data often encounter unseen distributions during operation, leading to performance degradation. TTA addresses this issue by allowing models to adapt their internal parameters using unlabeled test data, without requiring retraining~\cite{lim2023ttn}. Several MLaaS platforms (e.g., \textit{Azure}, \textit{IBM Personalizer}\footnote{\url{https://docs.aws.amazon.com/personalize/}}
) already support real-time behavior adaptation by leveraging live user feedback and streaming events. Traditional adaptation relies on service substitution, whereas TTA enables direct runtime adjustment within the composition without service replacement.

\textit{Traditional TTA techniques are designed for single-model settings and are not directly applicable to MLaaS composition~\cite{schneider2020improving}.} Incorporating TTA into MLaaS composition can alter both functional and QoS constraints, thereby affecting \textit{composability}, i.e., the ability of multiple services to interact correctly while satisfying end-to-end requirements~\cite{kanneganti2025adaptive}. Existing studies primarily rely on static functional and QoS attributes to determine composability~\cite{guidara2016dynamic,ibrahim2015semantic}. However, they do not account for the internal model changes introduced by TTA, limiting their ability to accurately determine composability after adaptation. In addition, TTA can introduce \textit{cross-service effects}, where adaptation in one service influences the behaviour and performance of other services within the composition~\cite{li2020federated}. Existing studies primarily examine these effects in standard MLaaS settings and do not explore how adaptation-induced changes affect composability in MLaaS. We identify the following key challenges in enabling personalized adaptation using TTA in MLaaS composition for IoT.

\vspace{-1mm}
\begin{itemize}[itemsep=0ex, leftmargin=2ex] 
\item \textbf{Personalized Adaptation:} Personalized adaptation can be applied at both the service and composition levels; however, service-level adaptation may alter internal behavior, potentially affecting composability with other services. Such changes can introduce bias toward specific input patterns and disrupt expected interactions within the composition. Existing approaches do not consider the impact of TTA-induced changes on composition prior to adaptation, and therefore cannot ensure that adapted services remain compatible. Moreover, service-level TTA can introduce unintended bias and domain shifts across services, affecting overall system performance~\cite{zhang2024enabling}. In contrast, applying TTA at the composition level may lead to unstable behavioural changes across services~\cite{iftee2025pfedbbn}. Therefore, an effective mechanism is required to monitor and regulate adaptation so that individual service contributions are preserved while maintaining overall composition stability.
 \end{itemize}

To address these challenges, we propose a novel Test-Time Adaptive (TTA) composition framework for MLaaS in IoT environments. First, we introduce a TTA-aware Composability Model (TCM) to determine whether adapted services remain composable within an MLaaS workflow. To achieve this, we design five adaptive composability rules that assess key adaptation and compatibility factors. The TCM ensures that local service adaptations do not adversely affect the compatibility and collaborative behavior of composed MLaaS. Building on this, a Service-level Adaptation Model (SAM) regulates personalized adaptation updates at the service level, preventing bias and domain shifts from degrading overall composition performance. Overall, the framework enables MLaaS compositions to adapt to data shifts without service substitution.

\vspace{-2mm}
\section{Related Work}
\subsection{Adaptive Service Composition}
MLaaS has emerged as an effective paradigm for integrating pre-trained models into IoT applications~\cite{ribeiro2015mlaas}. Due to diverse functional and QoS requirements, recent studies increasingly focused on MLaaS composition and federation to impove decision-making~\cite{xie2024skyml,wang2025collaborative,kourtellis2020flaas,perera2025reinforcement}. In dynamic IoT environments, adaptive composition becomes essential to maintain performance under evolving data distributions and system requirements~\cite{ibrahim2015semantic}. Existing adaptive composition approaches primarily rely on semantic matching, QoS attributes, behavioral planning, and contextual factors to support runtime substitution or re-composition~\cite{guidara2016dynamic,fujii2009semantics,wang2010adaptive}. For example, semantic-driven frameworks use ontology-based functional matching and QoS ranking for service replacement~\cite{ibrahim2015semantic}, while recent MLaaS studies adopt monitoring-driven re-composition to replace degraded services~\cite{kanneganti2025adaptive}. Although these techniques improve adaptability, they depend heavily on identifying suitable replacement services, which is often time-consuming and impractical. Moreover, existing composability techniques mainly rely on static semantic, QoS, contextual, or rule-based characteristics, with limited consideration of how internal model behaviour changes during adaptation~\cite{guidara2016dynamic,fujii2009semantics,wang2010adaptive}. In contrast, TTA introduces new composability challenges, as personalized adaptation may alter internal model characteristics and affect the compatibility, a problem that remains largely unexplored


\subsection{Test-Time Adaptation}
TTA has emerged as an effective solution in ML and FL for handling data and domain shifts without retraining. However, its application in adaptive MLaaS compositions remains largely unexplored. Standard TTA methods such as TTA-BN~\cite{schneider2020improving}, TTA-GRAD~\cite{wang2020tent}, and TTA-MEMO~\cite{zhang2022memo} update models through normalization alignment, entropy minimization, and gradient-based adaptation. In composed settings, uniformly integrating such updates can introduce bias and domain shift due to inter-service interactions. While recent works address this in traditional settings using layer-wise adaptation~\cite{sahoo2025layer, park2024layer} or similarity-based aggregation in FL~\cite{zhang2024enabling}, these strategies do not extend to MLaaS compositions where system-level dependencies must be considered. This highlights the need for controlled test-time adaptive composition at both service and composition levels to preserve overall stability.

\section{Key Definitions and Problem Statement}\label{MLaaS_problem}
This section introduces the key definitions for understanding the TTA in MLaaS composition problem.

\noindent\textbf{Definition 1 (MLaaS Service).} 
\textit{An MLaaS service $M$ is represented by the tuple}
\( M = \langle id, \mathcal{F}_m, QoS_m\rangle \)
\textit{where:}

\begin{itemize}[itemsep=0ex, leftmargin=2ex]

\item $id$ denotes the unique identifier of the MLaaS service.

\item $\mathcal{F}_m$ is the functional specification, encompassing the required model parameters including model weights ($w$), layer-wise gradient updates ($\Delta\theta$) representing weight ($w$), scale ($\gamma$) and shift ($\beta$) adjustments, and batch normalization (BN) parameters including scale ($\gamma$), shift ($\beta$), and BN statistics mean ($\mu_{BN}$) and variance ($\sigma^2_{BN}$).

\item $QoS_m$ denotes the \textit{evaluation metrics} such as accuracy, precision, recall, or other task-specific measures.

\end{itemize}

\noindent\textbf{Definition 2 (Adaptive MLaaS composition).} 
An MLaaS composition refers to the integration of multiple ML services, where their parameters or updates are aggregated to form a unified model for a shared task~\cite{wang2025collaborative, kanneganti2025adaptive}. In dynamic environments, compositions can adapt through service substitution or test-time adaptation to maintain system objectives. Let ${M_c}=\{M_1,\dots,M_n\}$ denote a set of services, where each service $M_i$ contributes parameters $\theta_i$. The composed model is defined as
\begin{equation}
\small
M_c^{t} = \sum_{i \in \mathcal{S}^{t}} \tilde{\theta}_i^{\,t}
\end{equation}
where $\mathcal{S}^{t}$ denotes the set of active services at adaptation step $t$, and $\tilde{\theta}_i^{\,t}$ represents the effective contribution of service $M_i$, which may evolve over time due to either replacement or internal adaptation.

\noindent\textbf{Problem Statement :} An MLaaS composition $M_c = \{M_1, \ldots, M_n\}$ is designed to jointly satisfy functional and QoS objectives (see \textit{Definition 1}), producing predictions over a continuous IoT data stream $X = \{x_t\}_{t=1}^{\infty}$. Over time, shifts in the input data distribution may degrade performance, necessitating adaptation. The problem is determining \textit{how} to apply TTA within $M_c$. In this setting, an underperforming service is first identified, and TTA is applied to that service, modifying its internal parameters (see \textit{Definition 1}) and producing an adapted service $M^*$. The key challenge is to determine whether $M^*$ remains composable with the performing services $M^p_c$ and can be integrated without disrupting inter-service dependencies within the composition. This is determined by a composability score function $\mathcal{CS}(M^p_c, M^*)$, defined as:
\begin{equation}
\small
\mathcal{CS}(M_c^{p}, M^*) = \begin{cases} 1, & \text{if } M^* \text{ aligns with } M^p_c \\ 0, & \text{otherwise} \end{cases}
\end{equation}
Once composability is determined, the adapted service must be integrated into the composition through a service-level TTA model that regulates its influence with respect to the performing composition $M_c^p$. This can be expressed as:
\begin{equation}
\small
\text{SAM}(M_j^*, M_c^p) \rightarrow \text{stable integration}
\end{equation}

\begin{figure}[t]
    \centering
    \includegraphics[width=\columnwidth]{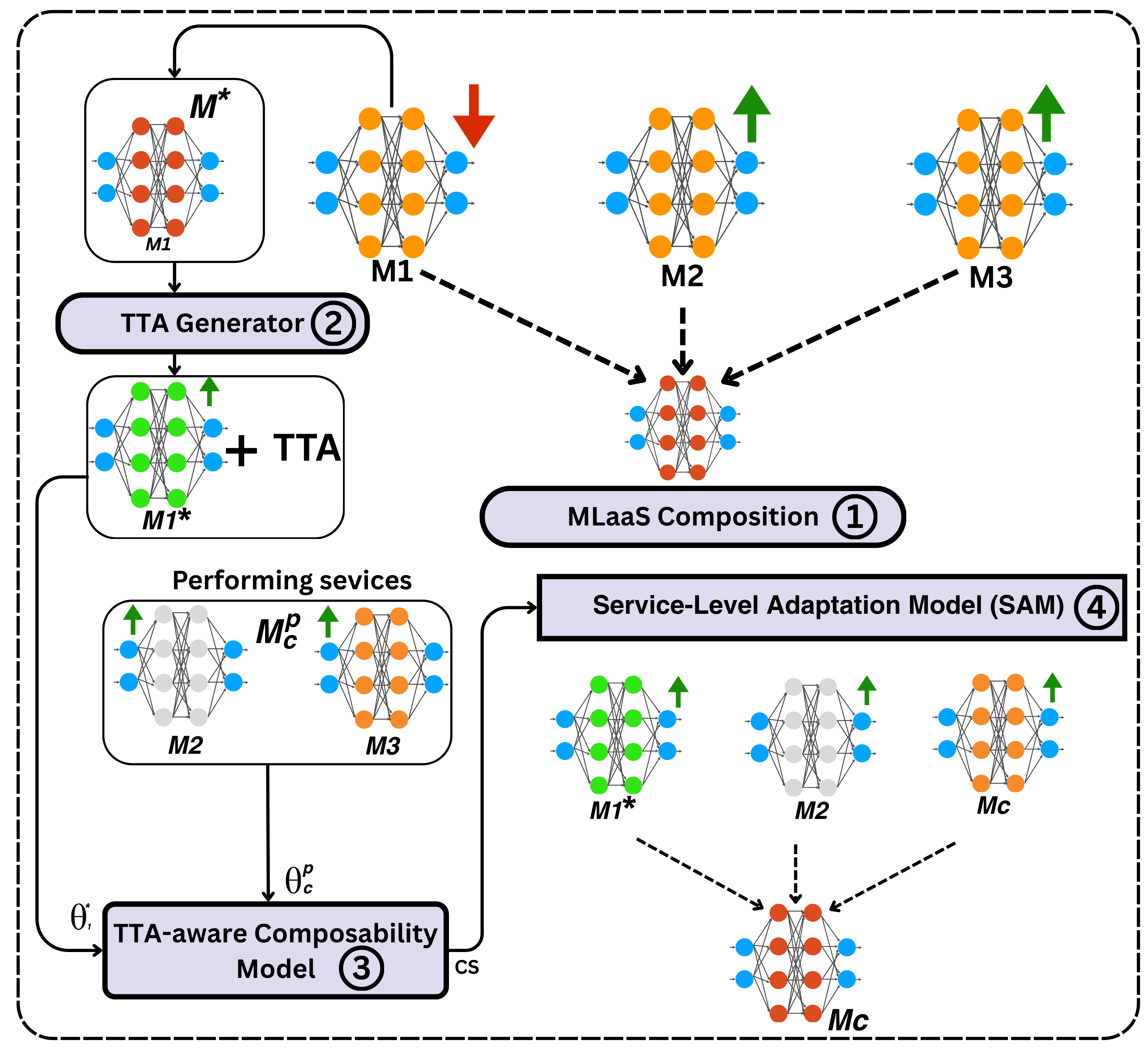}
    \caption{Proposed TTA MLaaS Composition Framework}
    \label{fig3}
    \vspace{-5mm}
\end{figure} 

\section{Test-Time Adaptation in MLaaS Composition for IoT Environments}
The proposed TTA-based MLaaS composition framework is illustrated in \textit{Fig. 1}. Once triggered, TTA is applied at both the service and composition levels. At the service level, TTA is applied to the underperforming service, and the TTA-aware composability model evaluates its compatibility with the remaining performing services. Based on this, the service-level adaptation model performs controlled adaptation while preserving interaction constraints. 

\vspace{-2mm}
\subsection{Test-Time Adaptability Rules}
We introduce Test-Time adaptability rules derived from TTA studies in ML and FL~\cite{sahoo2025layer,wang2020tent,zhang2022memo}. TTA modifies internal model parameters, including weights $(W)$, layer-wise updates $(\Delta \theta_l)$, batch normalization parameters $(\gamma, \beta)$, and statistics $(\mu, \sigma^2)$, influencing prediction probability distributions $(P)$. These rules compute a TTA-adaptability score in the range $(0,1)$ to determine service composability.



\textbf{a) Weight Drift Measurement (WDM).}
WDM evaluates the deviation of \textit{test-time adapted service} weight space from the performing composition. The divergence between model parameters is commonly used to determine whether an update remains consistent with participating models\cite{schneider2020improving}. We define the WDM between the test-time adapted service $M^{\ast}$ and the performing composition $M_c^{p}$ using cosine deviation as follows:
\begin{equation}
\small
\mathrm{WDM}_{M_c^{p},M^{\ast}}
=
1-
\frac{\langle W^{\ast}, W_c^{p}\rangle}
{\|W^{\ast}\|_2 \, \|W_c^{p}\|_2}
\end{equation}
where $W^{\ast}$ represents the model weights of the test-time adapted service $M^{\ast}$ and $W_c^{p}$ denotes the reference weight vector of the performing composition $M_c^{p}$. The value $\mathrm{WDM}$ quantifies the deviation introduced by TTA, a smaller values indicate that the adapted weights remain close to the composition, while larger values reflect stronger parameter drift.

\textbf{b) Update Consistency Score (UCS):} The UCS compares the layer-wise model parameter update vectors $\Delta\theta_l$ of the adapted service $M^{\ast}$ and the performing composition $M_c^{p}$. By measuring the directional similarity of these updates across corresponding layers, UCS evaluates whether $M^{\ast}$ remains consistent with the update behaviour of the performing composition using cosine deviation as follows:
\begin{equation}
\small
\mathrm{UCS}_{M_c^{p},M^{\ast}}
=
\sum_{l=1}^{L}
w_l
\frac{
\left\langle
\Delta\theta^{\ast}_{l},
\Delta\theta^{p}_{c,l}
\right\rangle
}{
\left\|\Delta\theta^{\ast}_{l}\right\|_2
\left\|\Delta\theta^{p}_{c,l}\right\|_2
+\epsilon
}
\end{equation}
where $\Delta\theta^{\ast}_{l}$ and $\Delta\theta^{p}_{c,l}$ denote the layer-wise gradient-based parameter update vectors of the adapted service and performing composition at layer $l$, respectively. Here, $L$ represents the total number of layers, $w_l$ denotes the contribution weight of layer $l$, and $\epsilon$ ensures numerical stability. Higher UCS values indicate greater update consistency.

\noindent\textbf{c) BN Update Alignment Score (BNUAS).}
BNUAS evaluates whether the batch normalization parameter updates introduced by the test-time adapted service $M^{\ast}$ remain directionally consistent with those of the performing composition $M_c^{p}$. TTA primarily modifies the batch normalization affine parameters  scale $(\gamma)$ and shift $(\beta)$. First, the cosine alignment of the layer-wise updates is computed, based on which the batch normalization update alignment score is defined.
\begin{equation}
\small
S_{\ell}
=
\frac{\langle \Delta\gamma_{\ell}^{\ast},\Delta\gamma_{\ell}^{p}\rangle}
{\|\Delta\gamma_{\ell}^{\ast}\|_2\|\Delta\gamma_{\ell}^{p}\|_2+\epsilon}
+
\frac{\langle \Delta\beta_{\ell}^{\ast},\Delta\beta_{\ell}^{p}\rangle}
{\|\Delta\beta_{\ell}^{\ast}\|_2\|\Delta\beta_{\ell}^{p}\|_2+\epsilon}
\end{equation}

\begin{equation}
\small
\mathrm{BNUAS}_{M_c^{p},M^{\ast}}
=
\sum_{\ell=1}^{L} w_{\ell} S_{\ell}
\end{equation}

where $\Delta\gamma_{\ell}$ and $\Delta\beta_{\ell}$ denote the layer-wise updates of the batch normalization scale $(\gamma)$ and shift $(\beta)$ parameters, $w_{\ell}$ is the contribution weight of layer $\ell$, $L$ is the number of batch normalization layers, and $\epsilon$ ensures numerical stability. Larger values indicate stronger update compatibility.

\noindent\textbf{d) BN Data Alignment Score (BNDAS).}
BNDAS measures how closely the BN statistics of the adapted service $M^{\ast}$ align with those of the performing composition $M_c^{p}$. Since BN mean and variance capture internal feature distributions, their deviation reflects distributional compatibility between the adapted service and the composition\cite{wang2020tent}. The BNDAS is computed by measuring the statistical distance as follows:
\begin{equation}
\small
D_{BN}(M_c^{p},M^{\ast})
=
\frac{1}{L}
\sum_{\ell=1}^{L}
\frac{
\|\mu_{\ell}^{\ast}-\mu_{\ell}^{p}\|_2
+
\|(\sigma_{\ell}^{2})^{\ast}-(\sigma_{\ell}^{2})^{p}\|_2
}{2}
\end{equation}
\begin{equation}
\small
\mathrm{BNDAS}_{M_c^{p},M^{\ast}}
=
\exp\!\left(-\frac{D_{BN}(M_c^{p},M^{\ast})}{\tau}\right)
\end{equation}

where $\mu_{\ell}$ and $\sigma_{\ell}^{2}$ denote the batch normalization mean and variance at layer $\ell$, $L$ is the number of batch normalization layers, and $\tau$ controls the sensitivity of the alignment score. A higher $\mathrm{BNDAS}_{M_c^{p},M^{\ast}}$ indicates stronger feature-distribution similarity between the adapted service and the performing composition, while lower values indicate a weaker alignment.

\textbf{e) Prediction Distribution Alignment Measurement (PDAM).}
PDAM evaluates whether the prediction behaviour of a TTA service remains consistent with the performing composition $M_c^{p}$. Deviations in class probability outputs may affect aggregation and prediction stability. To quantify this, we use the \textit{Wasserstein distance} ($WD$), which measures the discrepancy between probability distributions. The prediction distribution alignment between $M^{*}$ and $M_c^{p}$ is defined as:

\begin{equation}
\small
PDAM_{M_c^{p},M^{*}} = 1 - W_1(P^{*}, P_c^{p})
\end{equation}
Here, $P^{*}=(p_1^{*},\ldots,p_n^{*})$ denotes the prediction distribution of the adapted service, and $P_c^{p}=(p_1^{p},\ldots,p_n^{p})$ represents that of the performing composition. The WD $W_1(P^{*},P_c^{p})$ measures their divergence, where smaller values indicate stronger alignment and higher PDAM, reflecting better composability.

\begin{algorithm}[!t]
\caption{TTA-aware MLaaS Composability Model}
\label{alg2}
\begin{algorithmic}[1]
\State \textbf{Input:} $M_n$, $M_c^{p}$, $\mathcal{TTA}$, $\epsilon,\tau$
\State \textbf{Output:} $\mathbf{CS}$
\State $\mathbf{CS} \gets [\,]$

\For{each $t \in \mathcal{TTA}$}
    \State $M_n^{*,t} \gets \mathrm{ADAPT}(M_n, t)$
    \State Extract $W^{*,t}, \Delta\theta^{*,t}, P^{*,t}, \{\mu_\ell^{*,t}, (\sigma_\ell^2)^{*,t}, \Delta\gamma_\ell^{*,t}, \Delta\beta_\ell^{*,t}\}_{\ell=1}^{L}$
    \State $WDM \gets \exp\!\left(-\mathrm{WDM}(W^{*,t}, W_c^{p})\right)$
    \State $UCS \gets \mathrm{UCS}(\Delta\theta^{*,t}, \Delta\theta_c^{p}, \epsilon)$
    \State $BNUAS \gets 0,\; D_{BN} \gets 0$
    \For{each BN layer $\ell = 1$ to $L$}
        \State $S_{\ell} \gets \mathrm{BNUSim}(\Delta\gamma_{\ell}^{*,t}, \Delta\beta_{\ell}^{*,t}, \Delta\gamma_{\ell}^{p}, \Delta\beta_{\ell}^{p}, \epsilon)$
        \State $BNUAS \gets BNUAS + w_{\ell} S_{\ell}$
        \State $D_{BN} \gets D_{BN} + \mathrm{BNDist}(\mu_{\ell}^{*,t}, \mu_{\ell}^{p}, (\sigma_{\ell}^{2})^{*,t}, (\sigma_{\ell}^{2})^{p})$
    \EndFor

    \State $D_{BN} \gets \frac{1}{L}D_{BN}$
    \State $BNDAS \gets \exp\!\left(-\frac{D_{BN}}{\tau}\right)$
     \State $PDAM \gets \max(0,\min(1,\,1 - W_1(P^{*,t}, P_c^{p})))$
    \State $\mathrm{CS}_{M_n}^{t} \gets \frac{1}{5}(WDM + UCS + BNUAS + BNDAS + PDAM)$
    \State Append $(t,\mathrm{CS}_{M_n}^{t})$ to $\mathbf{CS}$
\EndFor

\State \Return $\mathbf{CS}$
\end{algorithmic}
\end{algorithm}


\subsection{TTA-aware MLaaS composability model}
We propose a TTA-aware MLaaS composability model to evaluate whether a test-time adapted service \(M^{\ast}\) remains composable with the performing composition \(M_c^{p}\). The Algorithm~\ref{alg2} outlines the composability evaluation process. The model evaluates compatibility using Test-Time Adaptability rules across multiple dimensions, including parameter drift and behavioral consistency. It takes as input the service, performing composition, and TTA-specific parameters(\textit{lines 1--4}). For each adapted service, the model computes composability indicators based on layer-wise BN parameter updates, model weights, and prediction behavior across different TTA techniques to assess compatibility (\textit{lines 5--18}). The composability score $CS_M^{t}$ is computed by combining these indicators. The score is appended for each TTA technique $t$, and the complete set of composability scores $CS$ is returned for decision making (\textit{lines 19--21}).

\begin{algorithm}[t]
\caption{Service-Level Adaptation Model}
\label{alg3}
\small
\begin{algorithmic}[1]
\State \textbf{Input:} $M_c^p$, $M$, $D_p=\{x_k\}_{k=1}^{N}$, $\mathcal{TTA}$, $\beta_s$
\State \textbf{Output:} $M_c^{new}$
\State $\theta_c^p \gets \{W_c^p, \mu_c^p, (\sigma_c^p)^2, \gamma_c^p, \beta_c^p\}$
\State $pw_c \gets 0$, \quad $pw^\ast \gets 0$
\For{each $t \in \mathcal{TTA}$}
    \State $M^{\ast,t} \gets \textsc{Adapt}(M, D_p, t)$
    \State $\theta^{\ast,t} \gets \textsc{Select}(M^{\ast,t}, t)$
    \State $\theta_c^{p,t} \gets \textsc{Select}(M_c^p, t)$
    \State $d_t \gets \|\theta_c^{p,t} - \theta^{\ast,t}\|_2$
    \State $s_t \gets \exp(-\beta_s d_t)$
    \State $pw_c^t \gets \dfrac{1}{1+s_t}$, \quad $pw^{\ast,t} \gets \dfrac{s_t}{1+s_t}$
    \State $pw_c \gets pw_c + pw_c^t$
    \State $pw^\ast \gets pw^\ast + pw^{\ast,t}$
\EndFor
\State $t^\ast \gets \arg\max_{t \in \mathcal{TTA}} pw^{\ast,t}$
\State $\theta^\ast \gets \theta^{\ast,t^\ast}$
\State $pw_c \gets \dfrac{pw_c}{pw_c + pw^\ast}$, \quad $pw^\ast \gets \dfrac{pw^\ast}{pw_c + pw^\ast}$
\State $\theta_c^{new} \gets pw_c \cdot \theta_c^p + pw^\ast \cdot \theta^\ast$
\State $\textsc{UpdateModelParameters}(M_c^{new}, \theta_c^{new})$
\State \Return $M_c^{new}$
\end{algorithmic}
\end{algorithm}

\vspace{-2mm}
\subsection{Service-Level Adaptation Model}
We propose a SAM to enable controlled integration at the service level within a performing composition $M_c^p$, as illustrated in Algorithm~\ref{alg3}. The model takes as input the performing composition, the target service $M$, the incoming test data $D_p$, and a set of TTA techniques $\mathcal{TTA}$ along with the scaling parameter $\beta_s$ (lines 1--3). The internal parameters of the performing composition are represented as $\theta_c^p = \{W_c^p, \mu_c^p, (\sigma_c^p)^2, \gamma_c^p, \beta_c^p\}$. \newline

\begin{table*}[t]
\centering
\footnotesize
\caption{Efficiency of the TCM model across corrupted datasets using TTA-BN, TTA-GRAD, and TTA-MEMO}\label{tab2}
\renewcommand{\arraystretch}{1.06}
\setlength{\tabcolsep}{3pt}
\begin{tabular}{l|cc|cc|cc|cc|cc|cc|cc|cc|cc}
\hline
\multirow{2}{*}{Method}
& \multicolumn{6}{c|}{MNIST-10C}
& \multicolumn{6}{c|}{CIFAR-10C}
& \multicolumn{6}{c}{CIFAR-100C} \\

\cline{2-19}

& \multicolumn{2}{c|}{\textit{TTA-BN}}
& \multicolumn{2}{c|}{\textit{TTA-GRAD}}
& \multicolumn{2}{c|}{\textit{TTA-MEMO}}
& \multicolumn{2}{c|}{\textit{TTA-BN}}
& \multicolumn{2}{c|}{\textit{TTA-GRAD}}
& \multicolumn{2}{c|}{\textit{TTA-MEMO}}
& \multicolumn{2}{c|}{\textit{TTA-BN}}
& \multicolumn{2}{c|}{\textit{TTA-GRAD}}
& \multicolumn{2}{c}{\textit{TTA-MEMO}} \\

\cline{2-19}

& Acc & Prec
& Acc & Prec
& Acc & Prec
& Acc & Prec
& Acc & Prec
& Acc & Prec
& Acc & Prec
& Acc & Prec
& Acc & Prec \\

\hline\hline

Rule based~\cite{wang2010adaptive}
& 44.90 & 43.10
& 48.65 & 47.53
& 50.70 & 49.26
& 41.13 & 70.66
& 38.97 & 28.55
& 24.82 & 57.28
& 35.64 & 54.21
& 35.63 & 41.48
& 50.86 & 64.27 \\

Similarity based~\cite{ibrahim2015semantic}
& 45.03 & 43.76
& 45.76 & 44.69
& 45.72 & 44.76
& 51.94 & 74.87
& 45.93 & 49.69
& 50.46 & 74.33
& 51.15 & 62.92
& 41.24 & 46.01
& 45.46 & 59.76 \\

Semantic based ~\cite{fujii2009semantics}
& 53.54 & 52.11
& 53.25 & 51.74
& 52.47 & 50.95
& 68.74 & 83.08
& 56.11 & 59.55
& 70.78 & 81.60
& 73.77 & 78.32
& 61.15 & 65.67
& 75.35 & 81.46 \\

QoS based~\cite{guidara2016dynamic}
& 48.54 & 46.37
& 47.79 & 45.90
& 47.99 & 46.60
& 34.45 & 64.51
& 35.25 & 37.84
& 33.37 & 64.01
& 38.12 & 55.47
& 33.21 & 31.02
& 35.03 & 53.43 \\

MLaaS based~\cite{kanneganti2025adaptive}
&64.28 &75.92
&63.62 &71.35
&62.74 &71.19
&63.13	&75.83
&56.15&	74.49 
&56.59	&74.40
&59.02&65.51
&54.74&63.92 
&56.88&64.80\\

\textbf{Ours} & \textbf{70.24} & \textbf{95.61} & \textbf{72.54} & \textbf{90.03} & \textbf{84.12} & \textbf{93.83} & \textbf{82.59} & \textbf{98.84} & \textbf{68.19} & \textbf{98.26} & \textbf{76.79} & \textbf{98.76} & \textbf{87.18} & \textbf{95.03} & \textbf{81.89} & \textbf{94.84} & \textbf{81.37} & \textbf{94.88} \\

\hline
\end{tabular}
\end{table*}

\begin{figure*}[t]
\centering

\includegraphics[width=0.31\textwidth]{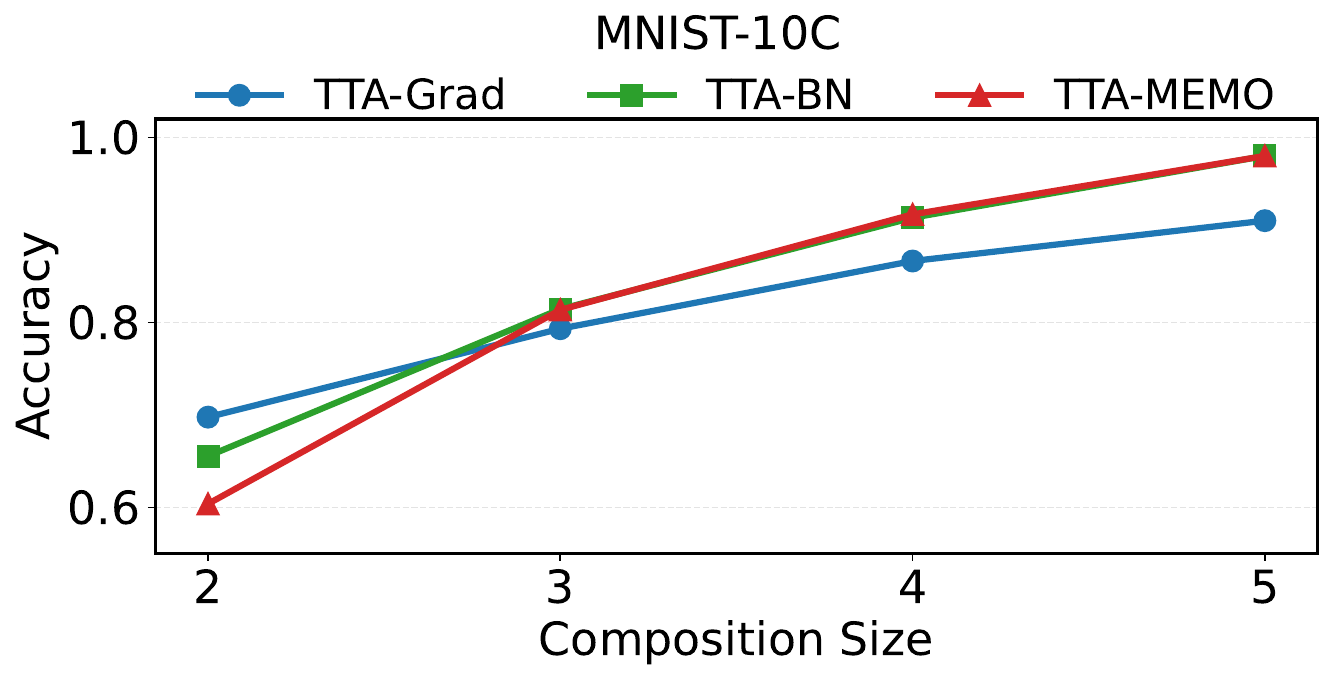}
\hfill
\includegraphics[width=0.31\textwidth]{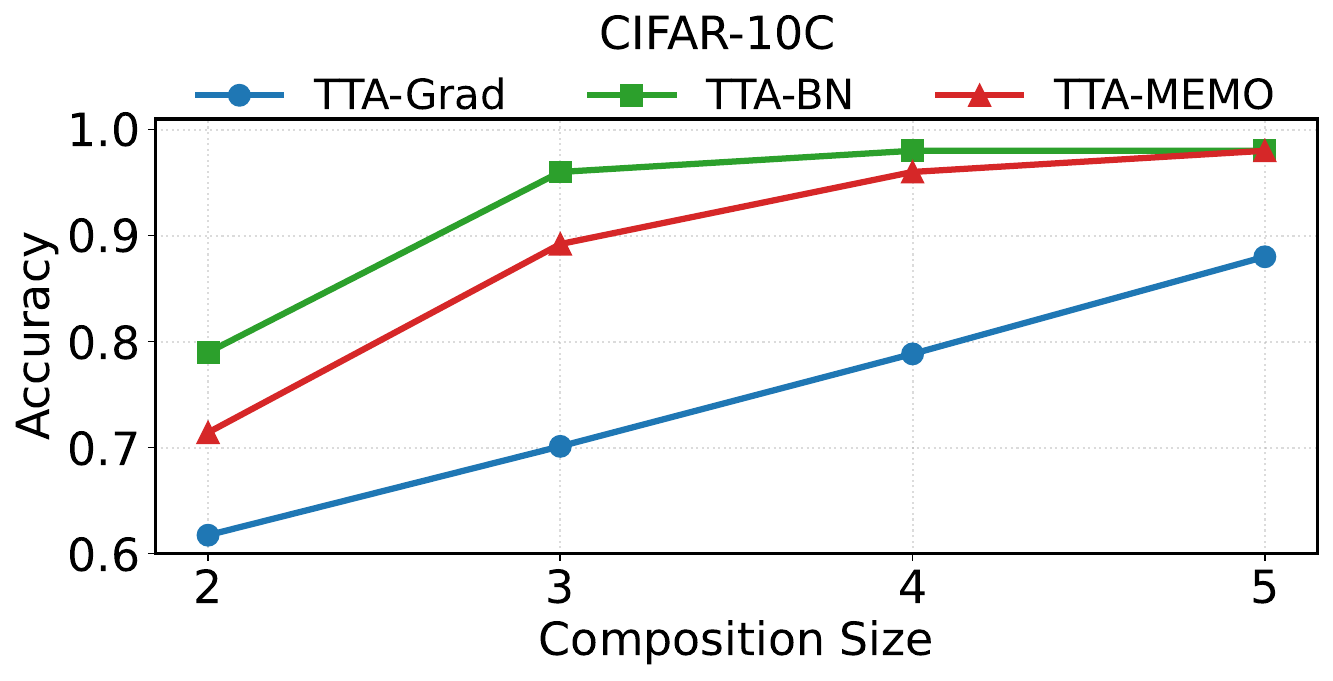}
\hfill
\includegraphics[width=0.31\textwidth]{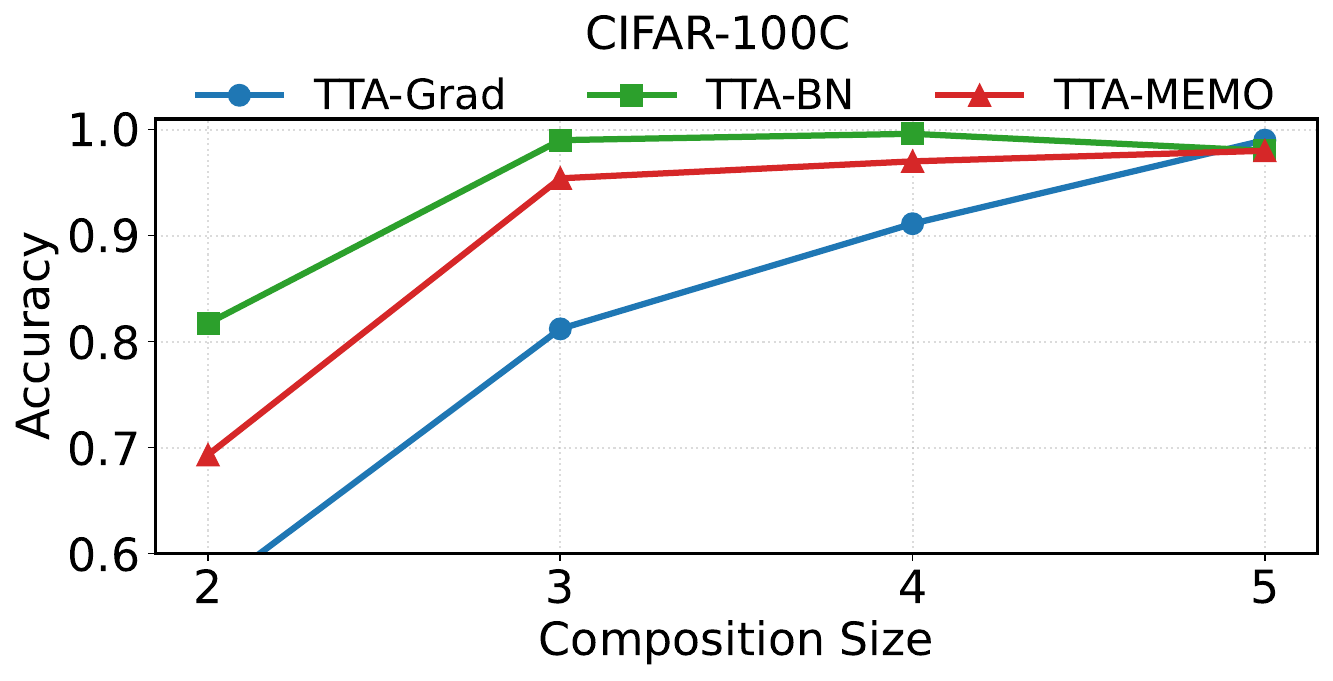}

\caption{Performance of the TCM under different composition sizes: (a) MNIST-10C, (b) CIFAR-10C, and (c) CIFAR-100C.}
\label{figexp2}

\end{figure*}
For each TTA technique, adaptation produces a technique-specific model $M^{\ast,t}$, from which
parameters $\theta^{\ast,t}$ are extracted and aligned with $\theta_c^{p,t}$ (lines 5--7). The deviation between them is computed using Euclidean distance (line 9) and transformed into a similarity score to derive preference weights (lines 10--11). These weights are accumulated across TTA techniques to capture overall consistency (lines 12--13). The best-aligned adaptation is selected (lines 15--16), and the final weights are normalized (line 17). The updated composition parameters are then obtained through weighted aggregation of $\theta_c^p$ and the selected adapted parameters $\theta^\ast$ (line 18), followed by updating the model parameters (line 19-20).

\section{Experiment Results and Discussion}\label{Experiment}
In this section, we evaluate the proposed TTA-based MLaaS composition framework for IoT environments. First,
we evaluate the TTA-aware MLaaS composability model across TTA-BN~\cite{schneider2020improving}, TTA-GRAD~\cite{wang2020tent}, and TTA-MEMO~\cite{zhang2022memo}, comparing it with baseline composability methods. Finally, we analyze performance gain and recovery ratio against traditional adaptive composition techniques across different corruption categories. All experiments were conducted on an Intel Core i7 machine with 16GB RAM using Python, and the source code and results are publicly available in the repository\footnote{\url{https://anonymous.4open.science/r/Personalised-TTA-Composition-223C}}.

\vspace{-2mm}
\subsection{Experiment Setup and Dataset}
To the best of our knowledge, no publicly available dataset specifically targets MLaaS services in IoT environments. Following recent MLaaS data generation frameworks~\cite{kanneganti2026machine}, we generated MLaaS composition data for our experiments. The evaluation is conducted using standard TTA benchmark datasets, including MNIST-C~\cite{mu2019mnist} (10,000 test images, 28$\times$28 grayscale, 10 classes), CIFAR10-C~\cite{hendrycks2019benchmarking} (10,000 test images, 32$\times$32 RGB, 10 classes), and CIFAR100-C~\cite{hendrycks2019benchmarking} (10,000 test images, 32$\times$32 RGB, 100 classes).

\subsection{Baseline}

We consider several baseline methods to evaluate different aspects of the proposed framework. First, to evaluate the efficiency of the proposed TCM, we compare against commonly used service selection and composition strategies, including rule-based methods, similarity-based techniques \cite{guidara2016dynamic}, semantic-based service matching \cite{ibrahim2015semantic}, and an MLaaS-based adaptive composition method \cite{kanneganti2025adaptive}. Next, to assess the performance gain and recovery behaviour of the proposed adaptation framework, we conduct experiments using adaptive composition baseline techniques such as context-based \cite{fujii2009semantics}, Semantic-based \cite{ibrahim2015semantic} and MLaaS-based \cite{kanneganti2025adaptive}
across different datasets. For a fair and consistent evaluation, the same evaluation metrics and thresholds were used across all baselines

\subsection{Experiment 1: Efficiency of the TTA-aware MLaaS Composability Model}
We investigate the efficiency of the proposed TCM against existing service composability techniques such as QoS attributes, semantic features, and MLaaS-based characteristics. Table~\ref{tab2} reports the composability accuracy across datasets. Traditional rule-based (41\%), similarity-based (47\%), and QoS-driven (39\%) methods show limited capability in capturing TTA-induced variations, while the semantic-based approach (63\%) performs comparatively better but remains inconsistent. MLaaS-based techniques improve performance but they do not fully capture effective compatibility due to their focus on functional specifications, overlooking parameter-level adaptations introduced by TTA. In contrast, TCM achieves consistently higher accuracy, with an overall average of approximately 85\%, including around 80\% (TTA-BN), 84\% (TTA-GRAD), and 89\% (TTA-MEMO), indicating improved reliability in composability decisions.
\textit{Fig.~\ref{figexp2}} illustrates the performance of TCM across varying service composition sizes (i.e., the length of service composition) under different TTA strategies. Smaller compositions show weaker performance due to limited service information, whereas larger compositions provide more reliable compatibility signals. As shown in \textit{Fig.~\ref{figexp2}(a)--(c)}, performance consistently improves with increasing composition size across MNIST-10C, CIFAR-10C, and CIFAR-100C, despite higher variability in more complex datasets.

\subsection{Experiment 2: Performance Gain and Computation time of SAM  Adaptation Under Benchmark Techniques}

\subsubsection{Performance Gain and Recovery}
In this section, we evaluate the performance gain of personalized adaptive composition using SAM across different corrupted datasets and adaptation settings. To simulate realistic distribution shifts, we consider both clean and mixed datasets, where the mixed dataset consists of 50\% clean samples and 50\% corrupted samples. The corrupted data is generated using multiple perturbation types, including brightness, blur, stripes, and zigzag patterns. We consider standard substitution techniques to replace underperforming services; however, identifying equivalent substitutes is not always feasible. To reflect this, we use services with 30\% and 50\% similarity, defined based on data-driven quality and performance.
Table~\ref{tab_SAM} summarizes the performance of SAM compared with substitution-based techniques. SAM maintains high accuracy under clean conditions, indicating stable behaviour without distribution shifts. Under mixed settings, all methods experience degradation due to data drift; however, SAM achieves consistent improvements (approximately 2--3\%) for \textit{glass\_blur}, \textit{stripe}, and \textit{zigzag}. In contrast, substitution-based methods show variable performance across scenarios, while remaining competitive in cases such as \textit{brightness}.

\begin{table}[t]
\centering
\footnotesize
\caption{Personalized adaptation comparison using SAM and substitution under different corruption settings.}
\label{tab_SAM}
\begin{tabular}{l|cc|cc|c}
\hline
\textbf{Corruption} 
& \multicolumn{2}{c|}{\textbf{Composition}} 
& \multicolumn{2}{c|}{\textbf{Substitution}} 
& \textbf{SAM} \\
\cline{2-6}
& \textbf{Clean} & \textbf{Mixed} 
& \textbf{30\%} & \textbf{50\%} 
& \textbf{Mixed} \\
\hline
brightness & 0.9972 & 0.6331 & \textit{0.8041} & \textit{0.8352} & \textbf{0.6545} \\
glass\_blur & 0.9923 & 0.9177 & 0.8992 & 0.8185 & \textbf{0.9308} \\
stripe & 0.9929 & 0.9312 & 0.8924 & 0.9076 & \textbf{0.9459} \\
zigzag & 0.9960 & 0.8993 & 0.8820 & 0.8600 & \textbf{0.9107} \\
\hline
\end{tabular}
\end{table}

\subsubsection{Computation Overhead}
Following the experimental setup in Table~III, we analyse the computation overhead of traditional substitution-based adaptive techniques and the proposed TTA-based adaptive composition framework. Traditional semantic-, context-, and MLaaS-based approaches rely on candidate discovery, attribute matching, validation, and iterative substitution, introducing significant overhead with execution times in the order of $10^6$--$10^7$. In contrast, the proposed SAM framework embeds composability assessment directly within adaptation, eliminating redundant substitutions and reducing computation time to the order of $10^3$--$10^4$. Experiments across 300--400 composition scenarios on MNIST and CIFAR datasets confirm up to $10^3\times$ lower computational overhead compared to substitution-driven methods, demonstrating efficient and practical adaptive composition.

\begin{table}[t]
\centering
\footnotesize
\caption{Overall computation time comparison with traditional adaptive techniques and the proposed test-time adaptive framework on MNIST and CIFAR datasets}
\label{tab:time_comparison_two_datasets}
\renewcommand{\arraystretch}{1.1}
\setlength{\tabcolsep}{6pt}
\begin{tabular}{l|cc}
\hline
\textbf{Method} & \textbf{MNIST} & \textbf{CIFAR} \\
\hline
Semantic-based~\cite{ibrahim2015semantic} & $6.3 \times 10^{6}$ & $1.0 \times 10^{7}$ \\
Context-based~\cite{fujii2009semantics} & $6.3 \times 10^{6}$ & $1.0 \times 10^{7}$ \\
MLaaS-based~\cite{kanneganti2025adaptive}  & $6.3 \times 10^{6}$ & $1.0 \times 10^{7}$ \\
Our     & $3.0 \times 10^{3}$ & $1.4 \times 10^{4}$ \\
\hline
\end{tabular}
\end{table}

\section{Discussion}\label{Experiment}
Our experiments demonstrate that the proposed TTA-based composition framework effectively reduces both computational cost and adaptation time. The framework ensures the compositions to remain stable and functional without opting for substitutions or re-composition. This highlights the practical advantage of TTA in dynamic IoT environments, where timely adaptation is critical and exhaustive re-composition is often infeasible. However, one key observation is that, while the framework is effective in retaining the performance of the composition, its ability to significantly improve predictive accuracy remains relatively limited. Addressing this issue requires a deeper investigation into the underlying service updates. Furthermore, the current design focuses on a single-step adaptation process, where TTA is applied in a one-pass manner to restore performance. Enabling incremental TTA could gradually improve model accuracy while avoiding instability and domain shift effects. however, exploring its role in continuous adaptation settings is left for future work. Additionally, the evaluation is limited to image classification datasets due to the scarcity of IoT benchmark datasets with TTA scenarios. Future work will explore sensor-based IoT datasets to assess the framework in more realistic IoT environments.\\

\section{Conclusion}\label{Experiment}

This paper proposes a novel Test-Time Adaptive (TTA) MLaaS composition framework for IoT environments. Enabling TTA in composed MLaaS introduces new challenges, as adaptation may alter service behaviour and affect composability. To address this, we propose a TTA-aware Composability Model (TCM) to determine the composability of adapted services. Experimental results show that TCM improves composability accuracy by about 19 percentage points over traditional approaches. Additionally, the service-level adaptation model mitigates bias and domain shift, achieving up to 2--3\% performance improvement under corrupted and mixed data conditions. Overall, the proposed framework provides an effective alternative to behaviour model substitution-based adaptation in dynamic IoT environments.

Future work will investigate incremental TTA mechanisms to enable gradual and stable improvements in model accuracy, addressing the limitations of fixed adaptation strategies in the current study while avoiding instability and domain shift effects.

\bibliographystyle{ieeetr}
\bibliography{Bibliography1}

\end{document}